\PassOptionsToPackage{table,dvipsnames}{xcolor}
\documentclass{article}
\usepackage{xcolor}

\usepackage[final]{corl_2026} % Uncomment for the camera-ready ``final'' version.
\title{Continual Robot Policy Learning
via \\Variational Neural Dynamics
}

\author{
\textbf{Jiaxu Xing} \hspace{0.7cm}
\textbf{Zhiyuan Zhu} \hspace{0.7cm}
\textbf{Yunfan Ren} \hspace{0.7cm}
\textbf{Ismail Geles}\\[0.9em]
\textbf{Yifan Zhai} \hspace{0.7cm}
\textbf{Rudolf Reiter} \hspace{0.7cm}
\textbf{Davide Scaramuzza}\\[1.6em]
Robotics and Perception Group,
University of Zurich
\vspace{-0.5cm}
}
\begin{document}
\maketitle
%==============================================================================
\begin{abstract}
Robots deployed in the real world rarely operate under a single fixed dynamics model: wind changes, payloads vary, batteries drain, contacts shift, and hardware wears.
Yet most learning-based controllers are trained once and deployed as if learning were complete.
This prevents the robot from using deployment experience to further improve task performance.
In this work, we propose a continual learning framework that uses real-world experience to improve robot policies under hidden and recurring dynamics.
Our method learns a condition-aware dynamics model from real state-action trajectories by combining an analytical physics prior with a neural residual for unmodeled effects.
A recurrent encoder infers the current hidden condition from recent interaction, and this estimate conditions both the residual model and the policy.
Policy learning is performed via differentiable simulation using diverse learned dynamics sampled from the latent model.
At deployment, these sampled conditions are replaced by conditions inferred online from recent real interaction, allowing the policy to recover recurring dynamics by recognition rather than residual re-fitting.
Through extensive simulation studies and real-world experiments, we demonstrate that the framework improves policy performance under diverse unobserved disturbances.
On real quadrotor trajectory tracking under changing wind, the policy recovers from recurring disturbances in roughly $1$\,s, about $5\times$ faster than online residual re-fitting.
It also reduces large-disturbance hover and tracking errors by $65.7\%$ and $53.3\%$ over the state-of-the-art online adaptation approaches.
Videos of the experiments are available at \url{https://youtu.be/t6NPMJzqOeE}.
\end{abstract}
\keywords{Continual Learning, Sim-to-Real, Differentiable Simulation}
\section{Introduction}

Learning-based control has achieved remarkable performance across robotics domains, spanning agile quadrotor flight~\cite{kaufmann23champion, geles2024demonstrating, xing2024multitask}, legged locomotion~\cite{miki2022learning, lee2020learning}, and visuomotor manipulation~\cite{chi2025diffusion}.
These control policies are typically trained in simulation, transferred to hardware, and deployed with fixed parameters.
This \emph{train-once, deploy-forever} paradigm has enabled impressive demonstrations, but it treats deployment as the end of learning rather than the source of the most relevant data.
Real-world dynamics are only partially specified before deployment and continue to vary afterward~\cite{aljalbout2025reality}.
In aerial robotics, external wind, payload changes, battery state, and hardware degradation all affect the closed-loop system.
The same issue arises in contact-rich control, where friction, contact compliance, and the unknown object mass can all alter the dynamics.
These factors are often hidden and non-stationary: a robot may encounter one condition, adapt to another, and later return to the first.
This makes one-shot adaptation inefficient: when a familiar condition reappears, the robot should recognize and reuse what it has already learned rather than adapt from scratch.
Robustness should therefore include identifying the active dynamics from interaction history, reusing prior knowledge, and continually improving behavior.
% FIG 2 - METHOD ARCHITECTURE (one-column or full-width)
% Place in methodology.tex, just after the "Two-stage approach" overview.
\begin{figure}[t]
    \centering
    \includegraphics[width=\linewidth]{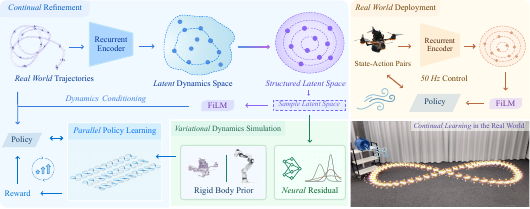}
    \caption{\textbf{Method overview.}
Our framework learns latent-conditioned residual dynamics from real trajectories, augmenting a rigid-body prior with a FiLM-modulated neural residual inferred from recent state-action history.
Sampled latents drive parallel differentiable policy training, while online inferred latents enable $50$\,Hz condition-aware deployment without privileged disturbance labels.
    }
    \label{fig:method_overview}
    \vspace{-0.35cm}
\end{figure}

Existing approaches only address parts of this problem.
Domain randomization~\cite{tobin2017domain} improves robustness by training over many simulated dynamics, but it usually produces a single policy that conservatively compromises across conditions.
Real2Sim2Real and residual-identification methods improve sample efficiency by leveraging real interactions to update a simulator or a residual dynamics model~\cite{kaufmann23champion, ren2026realworld, hwangbo2019learning, bauersfeld2021neurobem}. 
Recent differentiable-simulation methods push this further by fitting quadrotor residual dynamics from real flight data within seconds~\cite{pan2026learning, ren2026realworld}.
However, this adaptation is typically tied to the currently observed condition; when the dynamics change or a previous condition returns, the robot must relearn.
As a result, adaptation is limited by the dynamics distribution specified during simulation~\cite{wang2025environment}.

In this work, we propose a continual policy-learning framework that learns a condition-aware dynamics model from real trajectories and uses it to continually improve the policy.
The model starts from an analytical physics model and learns a neural residual for the unmodeled effects.
A recurrent encoder reads a short window of recent state-action history and outputs a latent embedding.
This embedding serves as a conditioning signal for both the residual dynamics model and policy training.
For policy training, we need to generate different hidden dynamics conditions in the simulation.
We therefore regularize the latent space so that samples from a simple prior produce realistic residual dynamics.
During training, sampled latents condition both the residual model and the policy in differentiable simulation; during deployment, they are replaced by latents inferred from recent real interaction.

Our main contributions are:
\begin{itemize}
    \item We develop a continual policy learning pipeline that connects real-world dynamics learning, differentiable policy optimization, and online deployment.
    The policy is trained under diverse learned dynamics and, at deployment, uses recent interaction to recover recurring conditions without re-fitting a new residual model.
    \item We introduce a Variational Neural Dynamics model that learns latent-conditioned residual dynamics from state-action trajectories while preserving an analytical robot dynamics prior.

    \item We validate the framework on real quadrotor tracking under changing wind and controlled simulation studies. 
    On hardware, the policy recovers from recurring disturbances in roughly $1$\,s, about $5\times$ faster than online residual re-fitting, and reduces large-disturbance hover and tracking errors by $65.7\%$ and $53.3\%$ over the state-of-the-art online residual adaptation baseline.
\end{itemize}
\section{Related Work}
\label{sec:related_work}
\textbf{Online Policy Adaptation in the Real World.}
A complementary line updates the policy or its dynamics model during deployment.
Classical adaptive control such as $\mathcal{L}_1$-MPC~\cite{hanover2021performance} offers fast disturbance rejection with stability guarantees but depends on predefined model structures.
Model-free RL methods~\cite{schulman2017proximal, schulman2015trust} support online updates in principle but require sample budgets unaffordable on physical robots~\cite{rudin2022learning}.
A recent line of work closes this gap via differentiable simulation: In~\cite{pan2026learning}, a residual dynamics model is fit from real flight and refines the policy in $\sim$5\,s, reducing hovering error by 81\% over $\mathcal{L}_1$-MPC, and in~\cite{ren2026realworld}, the authors extend this toward sustained performance growth (1.9 to 7.3\,m/s in $\sim$100\,s).
Both, however, fit a single residual to whichever condition is currently active and require fresh re-adaptation whenever conditions change.
Our work avoids this by training a single condition-aware policy across a distribution of conditions inferred from real flight data.

\textbf{Latent Conditioning on Unobserved Environmental Variables.}
History-based conditioning is widely used to adapt policies when dynamics are not directly observed.
Early methods used recent interaction for online system identification or implicit adaptation in randomized dynamics~\cite{yu2017preparing,peng2018sim}, while environment-probing and meta-RL approaches infer task or dynamics context from experience~\cite{rakelly2019efficient,o2022neural}.
In robotics, RMA~\cite{kumar2021rma,qi2023hand,kumar2022adapting} popularized history-inferred latent conditioning for legged locomotion, and DATT~\cite{huang2023datt} applied disturbance-aware conditioning to quadrotor tracking.
These works show that inferred context is an effective policy input under hidden dynamics, but the latent space is typically learned in simulation, often with privileged parameters or teacher policies, and the deployed policy is fixed.
Our work learns the latent dynamics representation from real state-action trajectories through residual dynamics prediction, then uses it to condition both the policy and a learned simulator for further policy optimization.

\section{Methodology}
\label{sec:methodology}
\paragraph{Continual Learning System.}
As illustrated in Figure~\ref{fig:method_overview}, our framework is an asynchronous loop that alternates between real-world interaction and differentiable policy improvement.
The loop begins with a base policy trained in a simplified simulator.
Its role is not to model the real dynamics perfectly, but to keep the robot near the task distribution so that useful trajectories can be collected safely.
We store these real state-action-transition data in a growing buffer $\mathcal{D}$ and use them to train a variational neural dynamics model.
The model keeps a known analytical dynamics prior and learns only a latent-conditioned neural residual for missing effects such as wind, payload changes, or hardware degradation.
This residual is conditioned on a low-dimensional latent variable $\mathbf{z}$, which is inferred from a short window of recent state-action history.
Intuitively, $\mathbf{z}$ represents the hidden condition currently affecting the robot.
The key design choice is that the same latent space is used in two ways.
At deployment time, the encoder infers $\mathbf{z}$ from recent real interaction, allowing the policy to adapt its behavior to the active condition.
After updating the latent dynamics model, we improve the policy by differentiating through simulated rollouts.
During policy training, we sample $\mathbf{z}$from a learned prior and use it as domain randomization within the differentiable simulator.
Thus, the latent is not only an adaptation feature for the policy; it is also a sampleable interface for generating condition-consistent dynamics during policy optimization.

\paragraph{Problem Formulation.}
We model the robot as a partially observed Markov decision process
$\mathcal{M}=(\mathcal{S},\mathcal{A},\mathcal{W},T,r,\gamma)$.
Here $\mathcal{S}$, $\mathcal{A}$, and $\mathcal{W}$ are the state, action, and hidden-condition spaces, $T$ is the transition kernel, $r$ is the reward, and $\gamma$ is the discount factor.
At time $t$, the robot observes its state $\mathbf{s}_t\in\mathcal{S}$ and applies an action $\mathbf{a}_t\in\mathcal{A}$, but the active condition $\mathbf{w}\in\mathcal{W}$ is hidden.
The transition depends on this hidden condition:
$
    \mathbf{s}_{t+1}
    \sim
    T(\cdot \mid \mathbf{s}_t,\mathbf{a}_t,\mathbf{w}).
$
For the quadrotor experiments, we use
$\mathbf{s}_t=(\mathbf{p}_t,\mathbf{q}_t,\mathbf{v}_t)\in\mathbb{R}^{10}$,
where $\mathbf{p}_t$ is position, $\mathbf{q}_t$ is orientation represented by a unit quaternion, and $\mathbf{v}_t$ is linear velocity.
The action is
$\mathbf{a}_t^\top=[T_t,\boldsymbol{\omega}_t^\top]\in\mathbb{R}^{4}$,
where $T_t$ is collective thrust and $\boldsymbol{\omega}_t$ is the commanded body rate~\cite{kaufmann2022benchmark}.
\begin{wraptable}[15]{R}{0.43\columnwidth}
\vspace{-1.2em}
\scriptsize
\setlength{\tabcolsep}{0pt}

\refstepcounter{algorithm}
\label{alg:continual_vnd}

\begin{minipage}{\linewidth}
\hrule
\vspace{0.25em}
\noindent\textbf{Algorithm~\thealgorithm: Continual Policy Learning via VND}
\vspace{0.25em}
\hrule
\vspace{0.35em}

\begin{algorithmic}
\Require $f_{\mathrm{prior}}$, buffer $\mathcal{D}$, prior $p(\mathbf{z})$
\State Init. $E_\phi,D_\psi,\pi_\theta$
\vspace{0.2em}

\Statex \textsc{Repeat during deployment}
\vspace{0.1em}

\AlgPhase{1. Collect and infer}
\State $\mathbf{z}_t \gets E_\phi(\mathcal{H}_{t-C:t})$
\State $\mathbf{a}_t \gets \pi_\theta({\mathbf{o}}_t,\mathbf{z}_t)$
\State $\mathcal{D}\gets \mathcal{D}\cup
(\mathbf{s}_t,\mathbf{a}_t,\mathbf{s}_{t+1})$

\vspace{0.25em}
\AlgPhase{2. Learn latent dynamics}
\State Update $(\phi,\psi)$ with
\Statex \hspace{0.7em}
$\mathcal{L}_{\mathrm{vnd}} = \mathcal{L}_{\mathrm{dyn}}
+ \lambda_{\mathrm{rec}}\mathcal{L}_{\mathrm{rec}}
+ \lambda_{\mathrm{mmd}}\,\mathrm{MMD}(q_{\phi},p)$

\vspace{0.25em}
\AlgPhase{3. Improve policy}
\State Freeze $E_\phi,D_\psi$; sample $\mathbf{z}\sim p(\mathbf{z})$
\State Roll out with
\Statex \hspace{0.7em}
$\mathbf{s}_{t+1}
=
f_{\mathrm{prior}}(\mathbf{s}_t,\mathbf{a}_t)
+
D_\psi(\mathbf{s}_t,\mathbf{a}_t,\mathbf{z})$
\State Update $\theta$ by BPTT on
\Statex \hspace{0.7em}
$J(\theta)
=
\mathbb{E}_{\mathbf{z},\mathbf{s}_0}
\sum_{h=0}^{H-1}
\gamma^h
r(\mathbf{s}_h,\pi_\theta(\mathbf{o}_h,\mathbf{z}),\mathbf{s}_{h+1})$
\end{algorithmic}
\vspace{0.3em}
\hrule
\end{minipage}
\vspace{-1.0em}
\end{wraptable}

Because $\mathbf{w}$ is hidden, the optimal action generally cannot be chosen from $\mathbf{s}_t$ alone.
The recent trajectory contains information about the active dynamics: for example, the same command may produce different accelerations depending on the wind direction.
We therefore summarize recent interaction history with a learned latent code.
Let
$
    \mathcal{H}_{t-C:t}
    =
    \{(\mathbf{s}_{i},\mathbf{a}_{i})\}_{i=t-C}^{t-1}
$
be a rolling context window of length $C$.
A history encoder maps this context to a low-dimensional latent:
$
    \mathbf{z}_t
    =
    E_{\phi}(\mathcal{H}_{t-C:t})
    \in \mathbb{R}^{d}.
$
The policy is then conditioned on both the current task observation and the inferred latent:
$
    \mathbf{a}_t
    =
    \pi_{\theta}({\mathbf{o}}_t,\mathbf{z}_t).
    \label{eq:latent_policy}
$
\paragraph{Variational Neural Dynamics Model.}
The latent code must be \emph{inferable} from recent interaction at deployment and \emph{sampleable} during policy learning.
Our model is designed to make these two requirements compatible in one latent space.
It couples a GRU history encoder $E_{\phi}$, which maps recent state-action history to the active condition code, with a latent-conditioned residual dynamics model $D_{\psi}$, which corrects a known analytical prior.
Given $\mathbf{z}_t=E_{\phi}(\mathcal{H}_{t-C:t})$, the next state is predicted by augmenting a known nominal dynamics model $f_{\mathrm{prior}}$ with a neural residual:
$
    \hat{\mathbf{s}}_{t+1}
    =
    f_{\mathrm{prior}}(\mathbf{s}_t,\mathbf{a}_t)
    +
    D_{\psi}(\mathbf{s}_t,\mathbf{a}_t,\mathbf{z}_t).
$
For the quadrotor, $f_{\mathrm{prior}}$ is a rigid-body model without the hidden disturbance, integrated with RK4.
The residual model is a shared-backbone MLP with position, velocity, and orientation heads,
$
    D_{\psi}(\mathbf{s}_t,\mathbf{a}_t,\mathbf{z}_t)
    =
    \left(
        \Delta \mathbf{p}_{\mathrm{res}},
        \Delta \mathbf{v}_{\mathrm{res}},
        \Delta \boldsymbol{\xi}_{\mathrm{res}}
    \right),
$
where $\Delta \boldsymbol{\xi}_{\mathrm{res}}\in\mathbb{R}^{3}$ is a rotation-vector correction composed with the nominal orientation prediction through the exponential map.
This keeps the known robot structure in $f_{\mathrm{prior}}$ while allowing the latent residual to capture condition-dependent effects such as wind, drag, payload changes, friction, or contact-induced deviations.
Because the residual is conditioned on $\mathbf{z}_t$, the same nominal transition can be corrected differently under different inferred dynamics, without fitting a separate model per environment.

\paragraph{Learning an Informative Latent Space.}
We train $E_{\phi}$ and $D_{\psi}$ jointly from dynamics prediction, without privileged labels such as wind direction, payload mass, friction coefficient, or hardware state.
The key idea is to make one inferred latent explain multiple transitions from the same trajectory context.
Let
$
    \mathbf{r}_t
    =
    \mathbf{s}_{t+1}
    \ominus
    f_{\mathrm{prior}}(\mathbf{s}_t,\mathbf{a}_t)
$
denote the residual target, where $\ominus$ denotes the appropriate state difference, including the rotation-vector difference for orientation.
The dynamics loss is
$
    \mathcal{L}_{\mathrm{dyn}}(\phi,\psi)
    =
    \frac{1}{B}
    \sum_{i=1}^{B}
    \ell_{\delta}
    \left(
        D_{\psi}(\mathbf{s}_i,\mathbf{a}_i,\mathbf{z}_{k(i)})
        -
        \mathbf{r}_i
    \right),
$
where $\ell_{\delta}$ is the Huber loss and $k(i)$ maps transition $i$ to the trajectory context that produced its latent.
Sharing $\mathbf{z}$ across transitions forces the encoder to capture the common dynamics behind them, making residual prediction a self-supervised signal for the hidden condition.

\paragraph{Sampleable Latent Space for Policy Learning.}
The dynamics loss makes $\mathbf{z}$ informative on encoded real trajectories: different hidden dynamics should require different residual corrections.
However, policy learning also requires sampling hidden dynamics in simulation before observing a trajectory.
This creates a mismatch: during deployment, latents come from the encoder, while during policy training, latents are sampled from a prior.
If encoded latents occupy an arbitrary region, prior samples may query the residual model off support and produce unrealistic dynamics.
We therefore align the aggregated encoder distribution over $N$ context windows,
$
    q_{\phi}(\mathbf{z})
    =
    \frac{1}{N}
    \sum_{n=1}^{N}
    \delta\!\left(\mathbf{z}-E_{\phi}(\mathcal{H}_{n,0:C})\right)
$
with the sampling prior $p(\mathbf{z})=\mathcal{N}(\mathbf{0},\mathbf{I})$ using maximum mean discrepancy (MMD)~\cite{gretton2012kernel}:
$
    \mathcal{L}_{\mathrm{mmd}}(q_{\phi},p)
    =
    \left\|
        \mu_k(q_{\phi})-\mu_k(p)
    \right\|_{\mathcal{H}_k}^{2},
$
where $\delta(\cdot)$ is a Dirac measure, $\mu_k(\cdot)$ is the kernel mean embedding induced by an RBF kernel, and $\mathcal{H}_k$ is the corresponding RKHS.
The Gaussian prior is only a convenient sampling space, not a physical disturbance model.

Unlike a VAE, the objective has no per-sample KL term, ELBO, or trajectory reconstruction likelihood.
The latent is learned through residual dynamics prediction, while MMD only matches the aggregated set of encoded latents to the sampling prior.
MMD alone does not make the latent identifiable or assign semantics to coordinates; the dynamics loss prevents collapsed codes by requiring incompatible residuals to be explained by different latents.
Also, in practice, we find that an annealed schedule on the MMD loss can keep the stability of the learned embedding space. 
Further details can be found in our ablation studies.
During policy learning, the encoder and dynamics model are frozen and all contexts in the replay buffer are re-encoded when updating the latent model, which keeps old and new conditions in a common coordinate system and reduces latent drift across deployment rounds.
The full variational neural dynamics objective is
$\mathcal{L}_{\mathrm{vnd}} = \mathcal{L}_{\mathrm{dyn}}
+ \lambda_{\mathrm{rec}}\mathcal{L}_{\mathrm{rec}}
+ \lambda_{\mathrm{mmd}}\,\mathrm{MMD}(q_{\phi},p)$, 
where $\mathcal{L}_{\mathrm{rec}}$ is an auxiliary $\mathcal{L}_2$
reconstruction loss from the latent code to the normalized state-action context
window.  We found this empirically helpful for retaining trajectory information
and training the dynamics model with higher accuracy.

\paragraph{Policy Learning with Sampled Latent Dynamics.}
After learning the variational neural dynamics model, we improve the current policy by sampling hidden conditions from the learned latent prior rather than manually choosing randomized physical parameters.
At the beginning of each rollout, we sample
$
    \mathbf{z}
    \sim
    p(\mathbf{z})
    =
    \mathcal{N}(\mathbf{0},\mathbf{I}),
    \label{eq:latent_sampling}
$
and hold this latent fixed for the full episode.
The same latent is provided to the policy and to the dynamics model.
With $N_{\mathrm{env}}$ parallel environments, we sample independent latents
$\{\mathbf{z}^{(j)}\}_{j=1}^{N_{\mathrm{env}}}$ and assign one to each environment.
During BPTT, $\bar{\mathbf{o}}_t=\mathrm{norm}_{o}(\mathbf{o}_t)$ denotes the
normalized policy observation produced by the same task observation map
$O_{\mathrm{task}}$ used at deployment.
For a rollout horizon $H$, the policy produces $
    \mathbf{a}_t
    =
    \pi_{\theta}(\bar{\mathbf{o}}_t,\mathbf{z}),
    \label{eq:policy_action}
$
and the learned dynamics produce
$   \mathbf{s}_{t+1}
    =
    f_{\mathrm{prior}}(\mathbf{s}_t,\mathbf{a}_t)
    +
    D_{\psi}(\mathbf{s}_t,\mathbf{a}_t,\mathbf{z}).
$

The dynamics parameters $(\phi,\psi)$ are frozen, and the fixed sampled latent makes the rollout differentiable with respect to the policy parameters $\theta$.
Following differentiable policy learning~\cite{freeman2021brax,song2024learning}, we maximize
$
    \max_{\theta}
    \;
    \mathbb{E}_{\mathbf{z}\sim p(\mathbf{z}),\,
    \mathbf{s}_0\sim\rho_0}
    \left[
        \sum_{t=0}^{H-1}
        \gamma^t
        r(\mathbf{s}_t,\mathbf{a}_t,\mathbf{s}_{t+1})
    \right].
$
Here $r$ is the task reward, $\gamma$ is the discount factor, and $\rho_0$ is the initial-state distribution.
The expectation is estimated with parallel differentiable rollouts, and $\theta$ is updated with Adam~\cite{adam}.

\section{Experiments and Results}
\label{sec:experiments}
Our experiments are designed to evaluate whether the proposed framework enables continual policy learning in the real world.
Following the method pipeline, we first introduce the hardware setup and baselines, then present the experimental results, and perform an analysis of the learned latent space.

\paragraph{Setup and Baselines.}
Hardware experiments use an Agilicious-based~\cite{foehn2022agilicious} quadrotor tracking figure-eight references at $50$\,Hz under controlled wind, with motion capture state estimates and off-board execution of the encoder, policy, and asynchronous learning loop.
We evaluate hover and trajectory tracking under small and large disturbances, reporting position error over 5 repeated trials per setting.
In simulation, we also evaluate landing and perceptive hovering, and two additional manipulation tasks; the setup details are provided in the supplementary material.
Baselines cover the main alternatives: Base DiffSim, $\mathcal{L}_1$-MPC~\cite{hanover2021performance}, RMA~\cite{kumar2021rma}, DATT~\cite{huang2023datt}, and LOTF~\cite{pan2026learning}.
For a fair comparison, all learned methods use the same state and action interfaces, reward, training/evaluation splits, and disturbance settings; privileged quantities for RMA and DATT are used only during training and never at deployment.
Also, we include a ground truth baseline called "GT. Dist." 
This represents a policy trained exactly to the disturbance conditions during the test time, which serves as an approximate upper bound under the test scenarios.
Further details on the training hyperparameters and network architecture are available in the supplementary materials.

\subsection{Real-World Flight Results}
\label{sec:exp_real_world}

We evaluate our continual learning system in the setting that motivates our work: a real quadrotor repeatedly encountering hidden wind conditions.

\paragraph{Recovery under recurring wind.}
In this setting, we change the wind after the robot reaches steady tracking and measure how long it takes for the tracking error to return to its pre-switch level.
For a wind direction already represented in the latent distribution, the proposed policy recovers within roughly $1$\,s by inference alone: the encoder updates $\mathbf{z}$ from the recent trajectory window, and the fixed policy changes its feedback behavior accordingly.
The online residual baseline of~\cite{pan2026learning} must instead re-fit a residual model for the new condition and takes approximately $5$\,s to recover.
Across wind-shift directions, this gives a $\sim$$5\times$ improvement in recovery time and shows that recognizing a recurring condition is faster than relearning it.
For more details, we refer the readers to the supplementary materials.

\begin{wrapfigure}[29]{r}{0.58\textwidth}
    \centering
    \vspace{-10pt}
    \includegraphics[width=\linewidth]{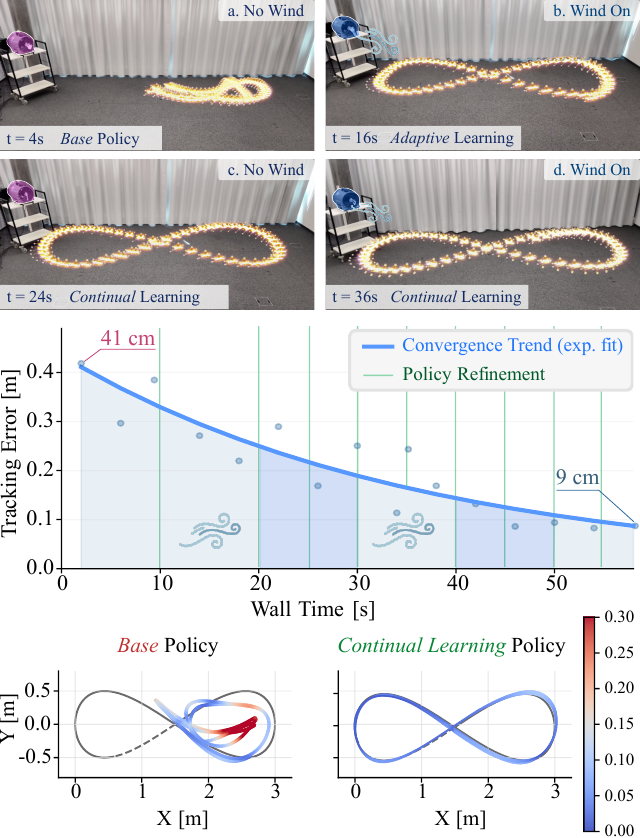}
    % \vspace{-10pt}
    \caption{\textbf{Real-world policy refinement.}
    On hardware, continual policy learning reduces figure-eight tracking error from $41$\,cm for the base policy to $9$\,cm after refinement.
    The lower panels show representative $xy$ tracking traces before and after refinement, illustrating improved trajectory tracking under the same deployment setup.
    }
    \label{fig:real_analysis}
\end{wrapfigure}

\paragraph{Revisiting a previous condition.}
We evaluate real-world policy refinement under intermittent wind disturbances.
Starting from the base policy, we repeatedly collect flight data and update the policy while the quadrotor tracks the same figure-eight trajectory.
Figure~\ref{fig:real_analysis} shows that tracking error decreases over wall-clock time, from about $41$\,cm initially to about $9$\,cm after refinement.
This demonstrates the continual-learning role of our framework: deployment data are not only used to fit a short-lived correction, but are converted into policy improvement that persists across subsequent flights.
The result shows that the learned dynamics model can support sustained policy refinement from real interaction under changing disturbance conditions.
\paragraph{Adaptation to Unmodeled Hardware Disturbances.}
We further evaluate an out-of-distribution hardware shift with a cut propeller, a setting that is difficult to anticipate through predefined domain randomization.
Because the actuation residual must be learned from real flight, this experiment tests whether the framework can incorporate genuinely new dynamics online.
Our method adapts within the first two refinement iterations (from $52$\,cm to $24$\,cm) and continues improving thereafter until $ 12$\,cm, showing that the same continual-learning loop can handle newly encountered residual dynamics beyond the nominal simulator.
Further details are provided in the Supplementary Material.

\subsection{System Analysis}
\paragraph{Dynamics Learning and Latent Space Analysis.}
We next examine whether the learned model captures the hidden varying dynamics rather than only improving control performance.

\begin{figure}[t]
    \centering
    \begin{minipage}{0.505\textwidth}
        \centering
        \includegraphics[width=0.99\textwidth]{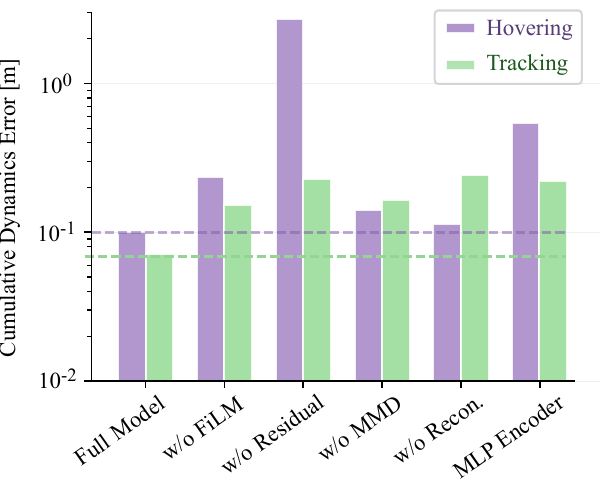}
\caption{\small \textbf{Dynamics-model ablations.}
The full model achieves the lowest cumulative rollout error on hovering and tracking, confirming the importance of latent-conditioned residual dynamics.}
        \label{fig:ablation_studies}
    \end{minipage}
    \hfill
    \begin{minipage}{0.485\textwidth}
        \centering
        \includegraphics[width=1.0\textwidth]{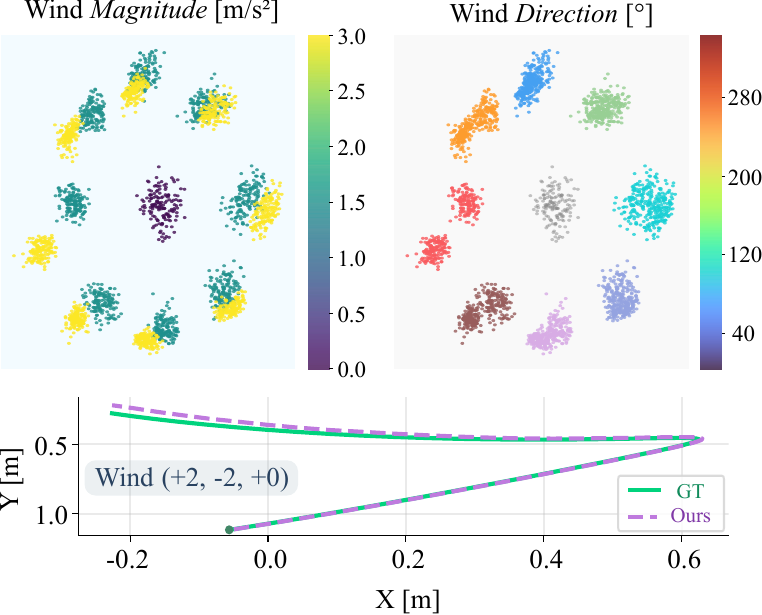}
\caption{\small \textbf{Latent dynamics analysis.}
Unsupervised latents cluster by wind direction and magnitude, and the learned dynamics accurately reproduce a representative wind-conditioned rollout.}
        \label{fig:latent_space}
    \end{minipage}
    \vspace{-0.3cm}
\end{figure}

Figure~\ref{fig:latent_space} shows encoded trajectory windows for different wind conditions in the simulation.
For the details on quadrotor models and wind simulation, we refer the readers to the Supplementary Materials.
As shown in Fig~\ref{fig:latent_space}, although wind labels are never used for training, the latent embeddings form clear clusters by wind direction and magnitude, indicating that the recurrent encoder discovers a structured dynamics representation from the state-action history.
The bottom plot in Fig~\ref{fig:latent_space} shows an open-loop rollout example, where the initial state and the action sequence are identical.
\begin{wraptable}[12]{R}{0.4\textwidth}
    \centering
    % \vspace{-1.4em}
    \scriptsize
    \setlength{\tabcolsep}{2pt}
    \renewcommand{\arraystretch}{1.25}
    \begin{tabular*}{\linewidth}{@{\extracolsep{\fill}} l cc cc @{}}
    \toprule
    \multirow{2}{*}{\textbf{Method}} 
        & \multicolumn{2}{c}{\textbf{Hover}}
        & \multicolumn{2}{c}{\textbf{Tracking}} \\
    \cmidrule(lr){2-3} \cmidrule(lr){4-5}
        & \textit{Small}$\downarrow$ & \textit{Large}$\downarrow$ 
        & \textit{Small}$\downarrow$ & \textit{Large}$\downarrow$ \\
    \midrule
    Base DiffSim  & 0.328 & 1.228 & 0.492 & 1.363 \\
    RMA  & 0.132 & 0.641 & 0.195 & 0.258 \\
    L1-MPC        & 0.134 & 0.552 & 0.121 & 0.281 \\
    DATT~\cite{huang2023datt}          & \cellcolor{Orange!10}0.009 & 0.231 & 0.082 & \textit{crash} \\
    LOTF~\cite{pan2026learning}         & \cellcolor{YellowGreen!25}{0.008} & \cellcolor{Orange!10}0.105 & \cellcolor{Orange!10}0.045 & \cellcolor{Orange!10}0.137 \\
    \cmidrule(lr){1-5}
    \rowcolor{cvprblue!8}
    \textbf{Ours} & \cellcolor{YellowGreen!25}{0.008} & \cellcolor{YellowGreen!25}{0.036} & \cellcolor{YellowGreen!25}{0.042} & \cellcolor{YellowGreen!25}{0.064} \\
    \bottomrule
    \end{tabular*}
    % \vspace{2pt}
    \small 
    \caption{\small \textbf{Benchmark Comparison}. Errors are in meters. Baselines follow \cite{pan2026learning}. Best in \colorbox{YellowGreen!25}{green}, second in \colorbox{Orange!10}{orange}.}
    \label{tab:drone_sim}
\end{wraptable}

The result shows that the latent-conditioned residual model accurately predicts condition-specific motion, matching the ground-truth trajectory below 0.1 m.
These results support the two requirements needed for policy learning: the hidden dynamics can be inferred from real trajectories, and the latent embedding can generate plausible condition-specific rollouts for training.

\paragraph{Comparison with State-of-the-Art Controllers.}
Table~\ref{tab:drone_sim} compares our method with fixed sim-trained control, latent adaptation, classical adaptive control, and online residual fitting.
Our method matches the best baseline on small-disturbance hovering and achieves the lowest error in the other three settings.
Compared with LOTF~\cite{pan2026learning}, the strongest online residual-adaptation baseline, our method reduces large-disturbance hover error from $0.105$\,m to $0.036$\,m and large-disturbance tracking error from $0.137$\,m to $0.064$\,m, corresponding to improvements of $65.7\%$ and $53.3\%$.
It also slightly improves small-disturbance tracking over LOTF, while avoiding the crash observed for DATT under large-disturbance tracking.
\paragraph{Ablation Studies.}
Figure~\ref{fig:ablation_studies} evaluates how each modeling choice affects multi-step dynamics prediction on hovering and trajectory tracking.
The full model gives the lowest cumulative error in both tasks.
The ``w/o Recon.'' label in the plot refers to removing the auxiliary
state-action context reconstruction term, while keeping the residual dynamics
prediction loss.  It is not a VAE-style trajectory likelihood.
The annealed MMD schedule is also a stability safeguard: applying prior alignment before the dynamics loss has separated conditions can collapse the representation, while omitting it lets encoded latents drift away from the sampling prior.
Removing FiLM conditioning, MMD alignment, or the reconstruction objective also increases error, indicating that the model needs both condition-specific residuals and a sampleable latent space.
Replacing the recurrent encoder with an MLP further degrades performance, especially in hovering, confirming that recent history is important for inferring the hidden dynamics.
Overall, the ablations show that the performance gain comes from the combination of residual dynamics, history-based condition inference, and latent-space regularization, rather than from model capacity alone.

\paragraph{Comparisons of Learning Pipelines.}
\begin{figure}[t]
    \centering
    \includegraphics[width=\linewidth]{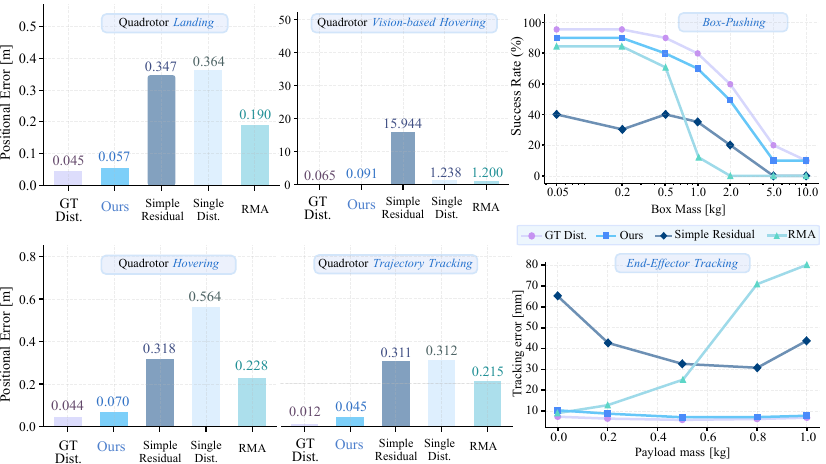}
    \caption{\textbf{Quantitative benchmark comparisons.}
    Across quadrotor landing, vision-based hovering, trajectory tracking, and contact-rich manipulation, the proposed latent dynamics framework approaches the ground-truth-disturbance oracle and outperforms residual, single-condition, and latent-adaptation baselines.
    Lower error is better except for box-pushing success rate.
    }
    \label{fig:quantitative_comparison}
    \vspace{-0.3cm}
\end{figure}

We next evaluate whether the learned dynamics distribution improves downstream policies, not only dynamics prediction.
Figure~\ref{fig:quantitative_comparison} compares our method with ground-truth disturbance conditioning, simple residual learning, single-disturbance training, and RMA-style latent adaptation across quadrotor and manipulation tasks.
For the detailed setup, we refer the readers to the Supplementary Materials.
Across quadrotor landing, hovering, and trajectory tracking, our method is consistently closest to the ground-truth oracle and substantially outperforms residual and single-condition baselines.
For example, in quadrotor trajectory tracking, our method reduces position error from $0.311$--$0.312$\,m for residual and single-disturbance baselines to $0.045$\,m, and improves over RMA from $0.215$\,m to $0.045$\,m.
The manipulation results show the same trend: in box pushing and end-effector tracking, our method remains close to the oracle across changing object or payload masses, while residual and latent-adaptation baselines degrade under larger hidden dynamics shifts.
These results indicate that sampling from the learned dynamics distribution exposes the policy to useful condition variation during training, leading to better generalization across both aerial and manipulation domains.

\section{Conclusion and Limitations}
We presented a continual policy learning framework for robots operating under hidden, changing dynamics.
Rather than treating deployment as the end of learning, our approach uses real interaction to learn a condition-aware dynamics model and continually improve the policy.
The framework combines an analytical physics prior with latent-conditioned neural residual dynamics, makes the learned dynamics distribution sampleable through prior alignment, and uses the same model for differentiable policy optimization and online condition inference.
Across real quadrotor flights and controlled simulation studies, the method learns structured dynamics representations, improves tracking under wind disturbances, adapts to unmodeled residual effects, and recovers recurring conditions without repeatedly re-fitting a new residual model.

\textit{Limitations and future directions.}
The current hardware experiments rely on accurate low-dimensional state estimates from motion capture.
In practice, onboard state estimates from visual-inertial odometry or SLAM will be noisy, delayed, and sometimes biased.
This matters because the same state estimates are used to learn the residual dynamics model; sensor noise can therefore be mistaken for unmodeled dynamics and lead to incorrect policy updates.
A possible direction is uncertainty-aware continual learning, where dynamics updates are weighted by state-estimation confidence and high-uncertainty transitions are filtered or modeled probabilistically.

\clearpage
% The acknowledgments are automatically included only in the final and preprint versions of the paper.
\acknowledgments{This work was supported by the European Union’s Horizon Europe Research and Innovation Programme under grant agreement No. 101120732 (AUTOASSESS) and the European Research Council (ERC) under grant agreement No. 864042 (AGILEFLIGHT).}
\bibliography{example}  % .bib
\clearpage
\appendix
\section{Supplementary Materials}
\label{app:supplementary}

Our supplementary materials provide the implementation details, experimental protocols,
baselines, and ablations supporting \emph{Continual Robot Policy Learning via
Variational Neural Dynamics}.  

\subsection{Overview}
\label{app:overview_map}
The supplementary is organized into the following  parts:
\begin{itemize}
    \item \textbf{Experimental Setup \& Environments:}
    \begin{itemize}
        \item Section~\ref{app:simulation_protocols} (\nameref{app:simulation_protocols}) outlines the details in various simulation experiments.
        \item Section~\ref{app:hardware_protocol} (\nameref{app:hardware_protocol}) introduces the real-world flight deployment infrastructure and details.
    \end{itemize}
    
    \item \textbf{Learning Objectives and Models:}
    \begin{itemize}
        \item Section~\ref{app:reward_details} (\nameref{app:reward_details}) defines the mathematical formulations of individual task rewards and hyperparameters.
        \item Section~\ref{app:vnd_details} (\nameref{app:vnd_details}) formalizes the Variational Neural Dynamics learning objectives and some training details.
        \item Section~\ref{app:policy_learning} (\nameref{app:policy_learning}) details the Backpropagation Through Time (BPTT) policy optimization objective.
    \end{itemize}

    \item \textbf{Implementation, Parameters, \& Benchmarks:}
    \begin{itemize}
        \item Section~\ref{app:training_compute} (\nameref{app:training_compute}) documents baseline convergence properties, wall-clock performance statistics, and hardware configurations.
        \item Section~\ref{app:metrics} (\nameref{app:metrics}) defines the quantitative evaluation metrics of the experiments.
        \item Section~\ref{app:state_action_history} (\nameref{app:state_action_history}) lists the underlying hyper-parameters utilized across models.
    \end{itemize}
\end{itemize}

\subsection{Simulation Experiment Setup}
\label{app:simulation_protocols}

All simulated tasks follow the same continual-learning template.
A nominal controller or current policy collects state-action transitions under hidden dynamics. 
The accumulated buffer trains the latent-conditioned residual dynamics model, and the policy is then optimized with sampled latents before evaluation on held-out hidden conditions. 
The tasks differ only in the robot model, hidden condition, and metric.

\paragraph{Quadrotor dynamics model.}
The quadrotor state follows the main-paper convention
$\mathbf{s}=(\mathbf{p},\mathbf{q},\mathbf{v})\in\mathbb{R}^{10}$, where
$\mathbf{q}$ is a unit quaternion.  We write $R(\mathbf{q})$ for the rotation
matrix induced by the quaternion when applying thrust in the world frame.  The
policy outputs collective thrust $T$ and commanded body rates
$\boldsymbol{\omega}$.  The differentiable policy-learning model is
\begin{equation}
    \dot{\mathbf{p}}=\mathbf{v}, \qquad
    \dot{\mathbf{v}}=\mathbf{g}+\frac{T}{m}R(\mathbf{q})\mathbf{e}_3, \qquad
    \mathbf{q}_{t+\Delta t}
    =
    \mathbf{q}_t\otimes
    \mathrm{Exp}_{\mathbb{H}}(\Delta t\,\boldsymbol{\omega}),
\end{equation}
where $\mathrm{Exp}_{\mathbb{H}}$ maps a rotation vector to a unit quaternion.
Position and velocity are integrated with RK4, while attitude is updated by
quaternion composition through the exponential map.  In wind experiments, the
hidden condition is implemented as a constant external acceleration in the
translational dynamics,
\begin{equation}
    \dot{\mathbf{v}}
    =
    \mathbf{g}+\mathbf{w}
    +
    \frac{T}{m}R(\mathbf{q})\mathbf{e}_3,
\end{equation}
with $\mathbf{w}$ fixed during an episode and unobserved by the deployed
policy.  The full-quadrotor configuration uses the Kolibri parameters: mass
$0.192$\,kg and per-motor thrust limit $3.5$\,N.  The arm offsets are
$(\pm0.041,\pm0.041,0)$\,m, the inertia is
$(1.42,1.65,2.13){\times}10^{-4}$\,kg\,m$^2$, and the motor time constant is
$0.0245$\,s.
\paragraph{Quadrotor wind model.}
To evaluate the system in controlled simulations, we use steady, hidden wind conditions that vary in direction and magnitude. 
The diagnostic data-generation scripts create 17 distinct wind conditions: 1 calm flight scenario and 8 horizontal (planar) wind directions tested at 2 speeds. 
Crucially, these wind labels are hidden from both the deployed flight policy and the self-supervised latent dynamics objective; they are utilized exclusively for simulation diagnostics and downstream evaluation plots.
\paragraph{Quadrotor Hovering.}
Hovering rollouts initialize the quadrotor around a fixed target with randomized
position, velocity, and attitude perturbations. 
The dynamics model predicts
conditioned position and velocity residuals.  
We report average position error under held-out wind conditions.

\paragraph{Quadrotor Trajectory tracking.}
Trajectory tracking uses a figure-eight reference. 
The data-generation controller randomizes initial states near the reference trajectory.
The residual model predicts position, velocity, and orientation residuals. 
We report mean tracking error against the reference.

\paragraph{Landing and perceptive hovering.}
The additional quadrotor simulation tasks use the same pipeline. 
Landing is reported by terminal position and success criteria.  
Perceptive hovering is evaluated using position error via the feature-based observation interface without relying on explicit state estimation, and a pretraining strategy is adopted from~\cite{pan2026learning}.

\paragraph{Manipulation tasks.}
The manipulation studies test whether the same loop extends beyond aerial robotics.  
The hidden condition is an unobserved object, payload, or contact
parameter that changes the closed-loop dynamics. 
Starting from a nominal analytical model and a stabilizing policy, the robot collects transitions, updates the latent-conditioned residual model, and improves the policy using rollouts sampled from the learned latent condition distribution.

\begin{figure}[t]
    \centering
    \includegraphics[width=0.94\linewidth]{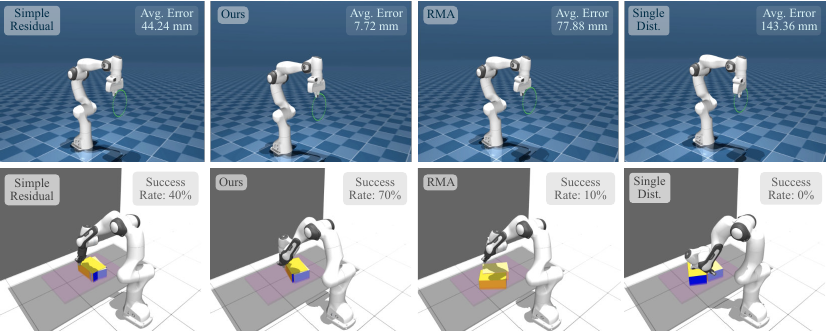}
    \caption{\textbf{Manipulation simulation setup.}
    We apply the same continual-learning pipeline to manipulation domains with
    hidden object, payload, and contact dynamics.  Deployment trajectories update
    a latent-conditioned residual model, and sampled latents expose the policy
    to recurring hidden manipulation conditions during differentiable policy
    optimization.}
    \label{fig:app_manipulation_simulation}
\end{figure}

In end-effector tracking, the hidden condition is a payload or object-mass
change that alters the effective arm dynamics. 
The policy observes only the
robot state and task reference. 
The encoder infers the active condition from
recent state-action history, while sampled latents provide data-driven domain
randomization over payload conditions.
We report Cartesian tracking error in
millimeters.

In planar box pushing, the hidden condition is contact and object dynamics,
including object mass and friction-induced deviations from the nominal model.
Because the same regime can recur across deployment episodes, a residual model that only fits the current episode must be rediscovered after each change. 
Our latent dynamics model can instead reuse a previously inferred condition through the history encoder.  
The metric is task success rate.

Across both manipulation settings, the key comparison is how the deployment data are
reused. 
Single-condition and simple residual baselines learn corrections tied
to one observed regime. 
Our method converts the accumulated interaction buffer into a sampleable distribution of hidden manipulation conditions, giving the
policy, a domain-randomization-like training interface learned from interaction
data rather than privileged physical parameters.

\subsection{Task Rewards}
\label{app:reward_details}

All methods within a task use the same reward.  At each control step the policy
outputs $u_t$, and $u_{\Delta t}=u_t-u_{t-1}$ denotes the action change.  The
robot state uses position $p$ (with height $z$), linear velocity $v$, body rate
$\omega$, linear acceleration $a$, and attitude
$R\in\mathrm{SO}(3)$.  We write $\Delta t$ for the control timestep.  The
per-step reward is the negative time-weighted cost,
$
    r = -\Delta t\, c ,
$
where $c$ is the task cost.  Each cost is a fixed, hand-tuned linear combination
of penalty terms $c_{(\cdot)}$.  Unless otherwise stated, every continuous term
is a smooth-$\ell_1$ (Huber) penalty
\begin{equation}
    \ell_\delta(e)
    =
    \begin{cases}
        \tfrac{1}{2} e^2, & |e| \le \delta,\\[2pt]
        \delta\big(|e| - \tfrac{1}{2}\delta\big), & |e| > \delta,
    \end{cases}
\end{equation}
applied to the relevant error $e$ (e.g. $c_v = \ell_\delta(\|v\|)$), keeping
gradients bounded for large errors while remaining quadratic near zero.

\paragraph{Quadrotor Hovering.}
The hovering cost penalizes target-height error
($c_{\text{height}}$), linear velocity ($c_v$), body rate ($c_\omega$), linear
acceleration ($c_a$), attitude deviation from upright ($c_R$), action deviation
from hover thrust ($c_u$), and collisions ($c_{\mathrm{collision}}$). 
In total,
\begin{equation}
    c_{\mathrm{hover}}
    =
    c_{\text{height}}
    +0.1\,c_v
    +0.15\,c_\omega
    +0.1\,c_a
    +0.1\,c_R
    +c_u
    +c_{\mathrm{collision}},
\end{equation}
with reward $r=-\Delta t\,c_{\mathrm{hover}}$.

\paragraph{Quadrotor Landing.}
Landing uses the hovering structure but replaces the constant-height objective
with a target landing position.  Thus $c_{\text{height}}$ becomes the
position-error term $c_p$, while all remaining terms and weights are unchanged:
\begin{equation}
    c_{\mathrm{land}}
    =
    c_{p}
    +0.1\,c_v
    +0.15\,c_\omega
    +0.1\,c_a
    +0.1\,c_R
    +c_u
    +c_{\mathrm{collision}}.
\end{equation}

\paragraph{Quadrotor Trajectory tracking.}
Trajectory tracking penalizes position, velocity, body-rate, acceleration, and
attitude errors relative to the current reference waypoint, together with action
smoothness and collision terms:
\begin{equation}
    c_{\mathrm{track}}
    =
    c_p
    +0.1\,c_v
    +0.01\,c_\omega
    +0.01\,c_a
    +c_R
    +0.15\,c_{\Delta u}
    +c_{\mathrm{collision}}.
\end{equation}
All reported tracking methods use the same reference, action limits, reset
distribution, and reward in simulation and hardware.
\paragraph{Franka end-effector tracking.}
Franka tracking uses Cartesian position error ($c_p$), orientation error
($c_R$), action regularization ($c_u$), and joint-limit penalty
($c_{\text{lim}}$) over a $5$\,s episode.  Continuous errors pass through
$\exp(-\beta e)$ with sharpness $\beta=5.0$.  The weights are $w_p=3.0$,
$w_R=0.3$, $w_{\dot q}=0$, $w_u=0.01$, and $w_{\text{lim}}=1.0$; the joint-limit
term activates within $0.05$\,rad of a limit.  These settings make Cartesian
tracking dominate the gradient while avoiding unnecessary damping of sustained
joint motion.

\paragraph{Box pushing.}
For box pushing, the PPO reward combines box--target progress, box orientation,
gripper--box proximity, side-contact shaping, joint-velocity terms, robot
target-posture, action regularization, success reward, and termination penalty:
\begin{equation}
    c_{\mathrm{box}}
    =
    c_{\text{prog}}
    +c_{\text{ori}}
    +c_{\text{prox}}
    +c_{\text{contact}}
    +c_{\dot q}
    +c_{\dot q\text{-lim}}
    +c_{\text{posture}}
    +c_{\Delta u}
    +c_{u}
    +c_{\text{term}}
    -r_{\text{succ}}.
\end{equation}
The logged setup uses $\Delta t=0.005$\,s, action repeat $4$, observation
history length $30$, action history length $5$, and delayed noisy observations
and actions during training.  Reward scales are
$w_{\text{prog}}=8.0$, $w_{\text{ori}}=6.0$, $w_{\text{prox}}=2.0$,
$w_{\text{contact}}=1.0$, $w_{\dot q}=1.0$,
$w_{\dot q\text{-lim}}=3.0$, $w_{\text{posture}}=0.75$,
$w_{\Delta u}=-0.1$, and $w_u=-0.1$.  
Success adds $500.0$ after $30$ consecutive success steps, with a $3.0$ success-wait reward; termination adds $-50.0$.  
The shaped terms use a $0.005$\,m box-target tolerance with $0.4$\,m margin, a $0.2$\,rad orientation tolerance with $\pi$ margin, and a $0.1$\,m gripper--box side-contact tolerance with $1.0$\,m margin.

\subsection{Hardware Setup}
\label{app:hardware_protocol}
The hardware setup matches the main real-world experiments. 
We use an
Agilicious-based quadrotor~\cite{foehn2022agilicious} tracking a figure-eight
reference at $50$\,Hz, with motion-capture state estimates and off-board
execution of the encoder, policy, and asynchronous learning process. 
Wind disturbances are controlled external conditions. 
Position error is reported over five repeated trials per setting with the same reference and evaluation window across methods.

\paragraph{Recurring-wind recovery.}
To evaluate the policy's adaptability, we conduct a wind-switch recovery experiment.
The quadrotor is initially flown under a constant wind condition until it achieves steady-state tracking. 
At time $t_{\text{switch}}$, the wind vector abruptly changes magnitude or direction. 

We define the \textit{Recovery Time} ($T_{\text{rec}}$) as the time required for the tracking error to return and remain within the pre-switch steady-state envelope. 
Formally, let $e(t) = \|x(t) - x_{\text{ref}}(t)\|_2$ denote the Euclidean tracking error at time $t$. 
Let $\epsilon$ define the tracking envelope, computed as the maximum error during the steady-state phase prior to $t_{\text{switch}}$. 
The recovery metric is defined as:
\begin{equation}
    T_{\text{rec}} = \inf \left\{ \tau \ge 0 \ \middle|\  e(t_{\text{switch}} + \tau + \Delta t) \le \epsilon \quad \forall \Delta t \ge 0 \right\}
\end{equation}
\begin{wrapfigure}[17]{r}{0.6\textwidth}
    \centering
    \includegraphics[width=\linewidth]{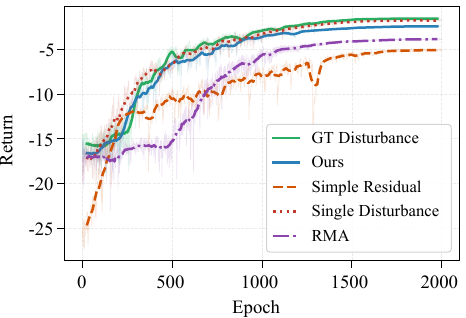}
    \caption{Training curves comparing return versus epoch across disturbance-modeling methods on the quadrotor task.}
    \label{fig:training_curve}
\end{wrapfigure}

\paragraph{Revisiting previous conditions.}
The revisit protocol exposes the robot to intermittent disturbances, including
sequences in which a previous wind condition returns. 
We measure the post-switch error spike, the time to return to nominal tracking, and the persistent tracking error after refinement.  
This separates condition recognition from continual policy improvement: a successful model should recover quickly when the condition is familiar and improve the policy over repeated deployment data.

\paragraph{Cut-propeller disturbance.}
The cut-propeller experiment tests a hardware disturbance not naturally covered
by hand-designed wind randomization.  
Because the actuator asymmetry is observed only through flight data, the experiment evaluates whether the continual loop
can absorb a genuinely new residual effect. 
The resulting performance improvement is
from $52$\,cm to $24$\,cm in the first two refinement iterations, followed by
continued improvement to $12$\,cm.

\subsection{Training Curves and Compute}
\label{app:training_compute}

Figure~\ref{fig:training_curve} shows a representative training curve for the
quadrotor trajectory-tracking task. 
We include it to document the optimization
behavior of one BPTT policy run; the evaluation metrics in the main paper are
computed separately on held-out disturbance conditions.

\paragraph{Training time and hardware.}
Policy training was run on a single NVIDIA RTX 4090 GPU.  Quadrotor hovering
and landing policies use $512$ parallel environments, $1500$ epochs, and
$250$ simulator steps per epoch, corresponding to $1.92{\times}10^8$
differentiable transition evaluations and $5$\,s rollouts at
$\Delta t=0.02$\,s.  The trajectory-tracking policy uses the same environment
count and epoch count with $125$-step rollouts, corresponding to
$9.6{\times}10^7$ transition evaluations and $2.5$\,s rollouts.  All three use
a policy MLP with hidden sizes $(512,512)$ and a four-dimensional
thrust/body-rate action.

Saved feature-hovering logs provide representative single-GPU timings:
$1000$ epochs with $512$ environments train in $130.6$\,s for the
simple-quadrotor wind run, $158.3$\,s for the full-quadrotor wind run, and
$188.5$\,s for the latent-conditioned feature-hovering run.  A Franka BPTT base
policy trains in about $10$ minutes, and a warm-started refined policy in about
$5$ minutes.  The logged box-pushing PPO policies use $8192$ parallel
environments and $1.8{\times}10^9$ simulator steps; one policy takes about
$2.8$--$2.9$ hours, with about $45$ seconds of JIT compilation before training.
The Franka residual model fit is shorter, about $15$--$20$ seconds for the
logged $50$k-transition datasets.

\begin{table}[t]
\centering
\small
\setlength{\tabcolsep}{4pt}
\renewcommand{\arraystretch}{1.15}
\begin{tabular}{p{0.24\linewidth} p{0.68\linewidth}}
\toprule
\textbf{Ablation} & \textbf{Question answered} \\
\midrule
\grayrow
w/o residual prior &
Does the model benefit from predicting only the correction to a physics prior,
rather than relearning the full transition? \\
w/o FiLM &
Is latent feature-wise modulation useful beyond simple concatenation of
$\mathbf{z}$ and the state-action vector? \\
\grayrow
w/o MMD &
Can the encoder learn predictive latents that are also safe to sample from
$\mathcal{N}(\mathbf{0},\mathbf{I})$ without aggregate prior alignment? \\
w/o reconstruction &
Does the auxiliary context reconstruction term help retain trajectory
information and improve multi-step prediction? \\
\grayrow
MLP encoder &
Is a temporal recurrent encoder necessary, or can a non-recurrent history
summary identify the active dynamics? \\
Latent-dimension sweep &
How sensitive is dynamics prediction and policy learning to the size of
$\mathbf{z}$? \\
\bottomrule
\end{tabular}
\vspace{0.2cm}
\caption{\textbf{Ablation map.} The full model combines a physics prior,
latent-conditioned residual dynamics, recurrent history encoding, context
regularization, and annealed MMD prior alignment.}
\label{tab:app_ablation_map}
\end{table}

\subsection{Metrics and Statistical Reporting}
\label{app:metrics}

For hardware experiments, the primary metric is position tracking error in
meters over repeated trials.  
Recovery time is reported in seconds and is
computed from the time at which the error returns to the pre-switch tracking
envelope. 
For simulation benchmarks, quadrotor tasks report mean position error, landing additionally reports terminal success criteria, end-effector tracking reports Cartesian tracking error, and box pushing reports success rate.  
When aggregate statistics are reported, each method uses the same seed set, disturbance split, and evaluation horizon within the corresponding task.

\subsection{State, Action, and History Representation}
\label{app:state_action_history}

For the quadrotor experiments, the physical state is
\begin{equation}
    \mathbf{s}_t =
    (\mathbf{p}_t,\mathbf{q}_t,\mathbf{v}_t)
    \in \mathbb{R}^{10},
\end{equation}
where $\mathbf{p}_t$ is position, $\mathbf{q}_t$ is a unit quaternion, and
$\mathbf{v}_t$ is linear velocity.  The action follows the main-paper
formulation,
\begin{equation}
    \mathbf{a}_t =
    (T_t,\omega_{x,t},\omega_{y,t},\omega_{z,t})
    \in \mathbb{R}^{4},
\end{equation}
where $T_t$ is collective thrust and the remaining entries are commanded body
rates.  The task observation additionally includes recent actions, allowing
command delay and smoothness to be represented.  Hovering observations contain
the current position, quaternion, velocity, and a short action history.
Trajectory-tracking observations use the same state-action interface and compute
the task error against the current reference waypoint.

The latent encoder consumes a rolling state-action context window
\begin{equation}
    \mathcal{H}_{t-C:t}
    =
    \{(\mathbf{s}_i,\mathbf{a}_i)\}_{i=t-C}^{t-1}.
\end{equation}
Unless otherwise stated, $C=20$ controller steps, corresponding to $0.4$\,s at
the $50$\,Hz hardware control rate.  The final configuration uses
$\mathbf{z}\in\mathbb{R}^{12}$; simulation-only diagnostic sweeps vary this
dimension to test sensitivity.  Input channels are normalized with statistics
computed from the training buffer, and the same normalization is reused across
our method and the history-conditioned baselines.
\begin{wraptable}[24]{r}{0.48\textwidth}
    \centering
    \vspace{0.4cm}
    \scriptsize
    \setlength{\tabcolsep}{2pt}
    \renewcommand{\arraystretch}{1.12}
    \begin{tabular*}{\linewidth}{@{\extracolsep{\fill}} lccc @{}}
    \toprule
    \textbf{Parameter} & \textbf{Quad. VND} & \textbf{Franka VND} & \textbf{Push PPO} \\
    \midrule
    State-action dim. & 13--14 & 21 & state hist. \\
    Residual output & 6--9 & 14 & policy only \\
    Context length & 20 & 20 & obs hist. 30 \\
    Latent dim. & 12 & 4 & -- \\
    Encoder & GRU & GRU & -- \\
    Encoder hidden & 128 & 256 & -- \\
    Residual hidden & 256 & 256 & -- \\
    Residual blocks & 2 & 3 & -- \\
    Conditioning & FiLM & FiLM & -- \\
    Classes & wind & 11 masses & -- \\
    VND epochs & per-task cfg. & 5000 & -- \\
    VND batch & per-task cfg. & 2048 & -- \\
    VND LR & per-task cfg. & $3{\times}10^{-4}$ & -- \\
    Policy hidden & 256, 256 & 256, 256 & 64$\times$4 \\
    Policy optimizer & Adam/BPTT & Adam/BPTT & PPO \\
    Policy epochs & per-task cfg. & 500 & -- \\
    Policy LR & per-task cfg. & $3{\times}10^{-3}$ & $6{\times}10^{-4}$ \\
    Horizon & per-task cfg. & 250 & 3000 \\
    Parallel envs & per-task cfg. & 256 & 8192 \\
    Control step & 0.02\,s & 0.02\,s & 0.005\,s \\
    \bottomrule
    \end{tabular*}
    \caption{\small \textbf{Model and training parameters.}
    ``Quad. VND'' lists the architecture shared by the quadrotor latent
    dynamics experiments; training-loop values are task specific. ``Franka
    VND'' comes from the manipulation tracking logs, and ``Push PPO'' lists the
    logged box-pushing policy settings.}
    \label{tab:app_model_parameters}
\end{wraptable}

\subsection{Variational Neural Dynamics Details}
\label{app:vnd_details}

\paragraph{Recurrent trajectory encoder.}
The encoder $E_\phi$ is a recurrent network.
Each state-action vector is first
projected through a dense layer with a GELU nonlinearity. 
A GRU aggregates the context window, and the final hidden state is projected to the latent code.  
The recurrent encoder is important because one transition often contains too little
evidence to identify the active disturbance; wind, for example, appears through accumulated changes in acceleration and attitude over multiple control steps.

\paragraph{Physics prior and residual head.}
The transition model keeps a known rigid-body prior and learns only the missing
condition-dependent component:
\begin{equation}
    \hat{\mathbf{s}}_{t+1}
    =
    f_{\mathrm{prior}}(\mathbf{s}_t,\mathbf{a}_t)
    +
    D_\psi(\mathbf{s}_t,\mathbf{a}_t,\mathbf{z}_t).
\end{equation}
For quadrotors, $f_{\mathrm{prior}}$ integrates nominal thrust and body-rate
dynamics with timestep $\Delta t=0.02$\,s.
The residual network predicts
corrections to position, velocity, and orientation.
Orientation residuals are
represented as rotation vectors and composed with the nominal orientation prediction through the exponential map, keeping the correction local on the rotation manifold.

\begin{figure}[t]
    \centering
    \includegraphics[width=\linewidth]{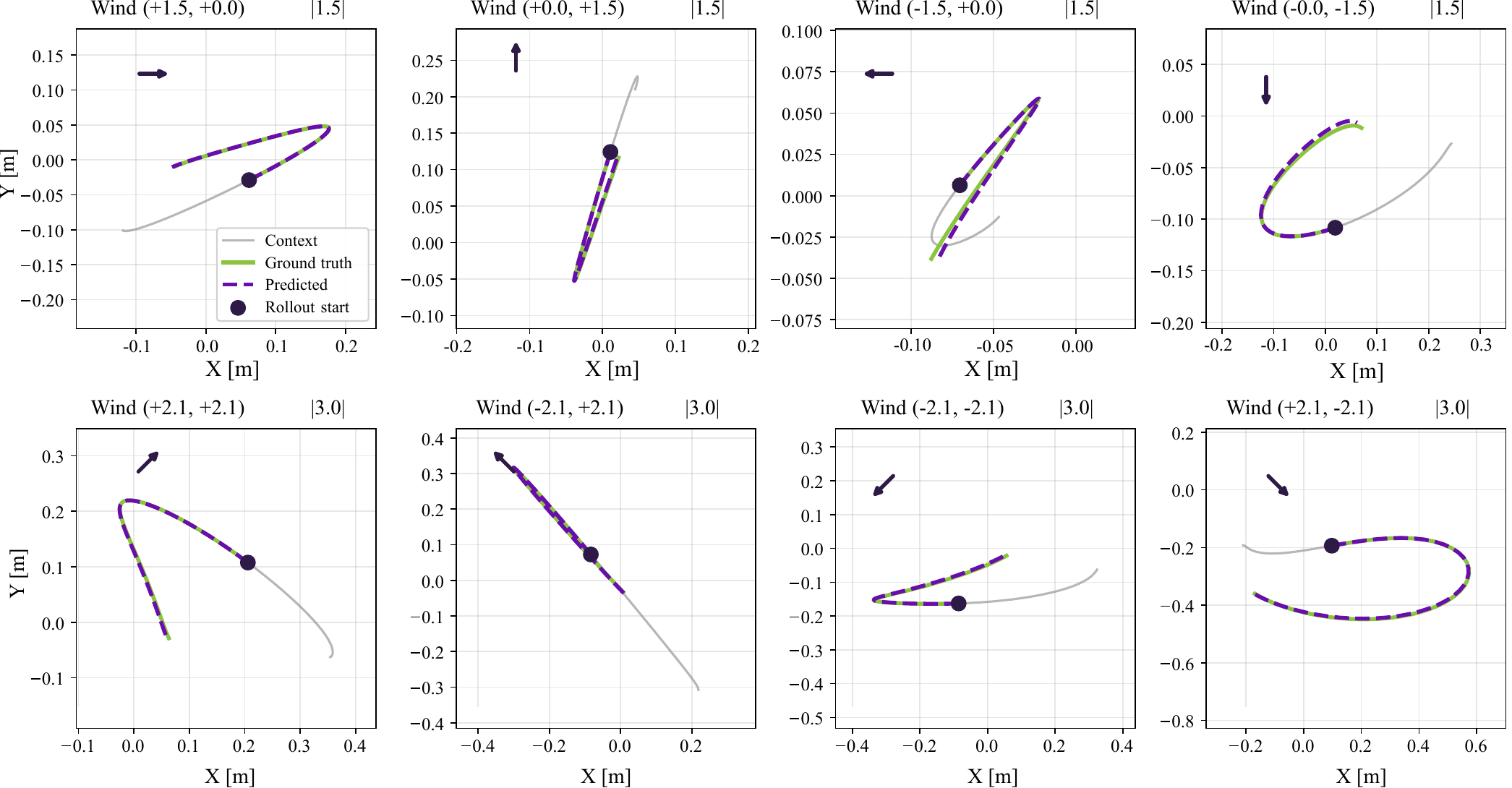}
    \caption{\textbf{Experiments on Learned Dyanmics.}
    Our learned Variational Dynamics Model is able to capture diverse disturbance conditions with very low open-loop cumulative errors.
    }
    \label{fig:dynamics_generaliztion}
    \vspace{-0.1cm}
\end{figure}

\begin{figure}[t]
    \centering
    \includegraphics[width=\linewidth]{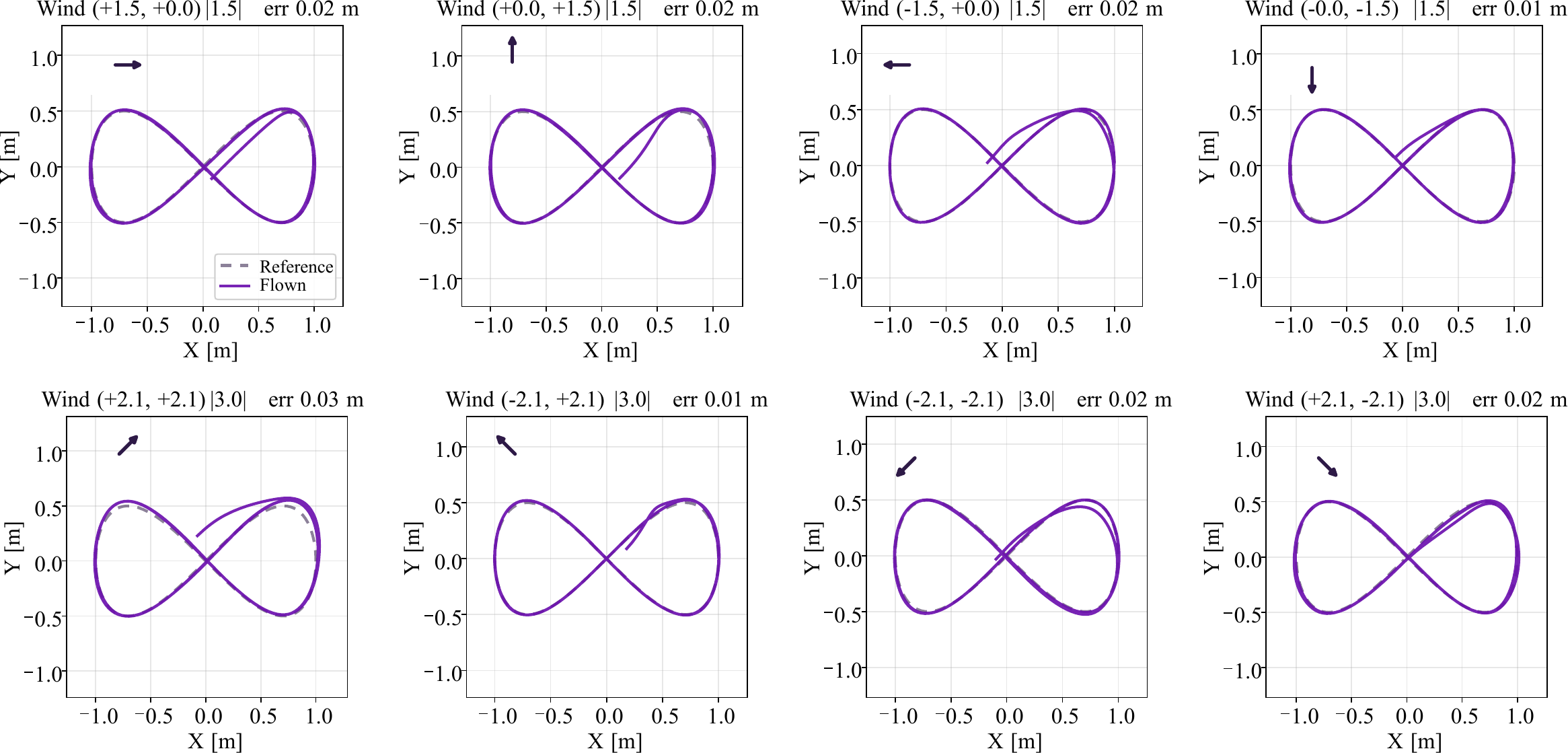}
    \caption{\textbf{Policy performance in diverse disturbance conditions.}
    Our learned policy through the continual learning framework is able to capture diverse external conditions.
    }
    \label{fig:policy_generaliztion}
    \vspace{-0.1cm}
\end{figure}

To evaluate the performance, we also include a rollout of the learned variational dynamics model under \textit{various} wind conditions, as shown in Fig.~\ref{fig:dynamics_generaliztion} and Fig.~\ref{fig:policy_generaliztion}.

\paragraph{Training objective.}
For a transition $(\mathbf{s}_i,\mathbf{a}_i,\mathbf{s}_{i+1})$ and a context
index $k(i)$, the residual target is
\begin{equation}
    \mathbf{r}_i =
    \mathbf{s}_{i+1}
    \ominus
    f_{\mathrm{prior}}(\mathbf{s}_i,\mathbf{a}_i),
\end{equation}
where $\ominus$ denotes Euclidean subtraction for position and velocity and a rotation-vector difference for orientation.
The dynamics loss is a Huber or
mean-squared residual-prediction loss over transitions:
\begin{equation}
    \mathcal{L}_{\mathrm{dyn}}
    =
    \frac{1}{B}
    \sum_{i=1}^{B}
    \ell\left(
        D_\psi(\mathbf{s}_i,\mathbf{a}_i,E_\phi(\mathcal{H}_{k(i)}))
        -
        \mathbf{r}_i
    \right).
\end{equation}
One inferred latent is shared by many transitions from the same context.  This
forces the code to explain the common hidden dynamics rather than memorize a
single transition.

The full objective used in the main paper is
\begin{equation}
    \mathcal{L}_{\mathrm{vnd}}
    =
    \mathcal{L}_{\mathrm{dyn}}
    +
    \lambda_{\mathrm{rec}}\mathcal{L}_{\mathrm{rec}}
    +
    \lambda_{\mathrm{mmd}}\mathcal{L}_{\mathrm{mmd}}.
\end{equation}
The reconstruction term decodes the latent back to the normalized context
sequence,
\begin{equation}
    \mathcal{L}_{\mathrm{rec}}
    =
    \frac{1}{N}
    \sum_{n=1}^{N}
    \left\|
        G_\eta(E_\phi(\mathcal{H}_n))
        -
        \mathrm{norm}(\mathcal{H}_n)
    \right\|_2^2 ,
\end{equation}
where $G_\eta$ is the auxiliary decoder and $\mathrm{norm}(\cdot)$ applies the training-buffer input normalization.
This term is an auxiliary context
regularizer, not a VAE likelihood, and it does not introduce a per-sample KL term.

The MMD term aligns the aggregated encoder distribution $q_\phi(\mathbf{z})$
with the sampling prior $p(\mathbf{z})=\mathcal{N}(\mathbf{0},\mathbf{I})$.  For
encoded latents $\{\mathbf{z}_i\}_{i=1}^{N}$ and prior samples
$\{\tilde{\mathbf{z}}_j\}_{j=1}^{N}$, we use the biased finite-sample estimator
\begin{equation}
    \mathcal{L}_{\mathrm{mmd}}
    =
    \frac{1}{N^2}\sum_{i,j} k(\mathbf{z}_i,\mathbf{z}_j)
    +
    \frac{1}{N^2}\sum_{i,j} k(\tilde{\mathbf{z}}_i,\tilde{\mathbf{z}}_j)
    -
    \frac{2}{N^2}\sum_{i,j} k(\mathbf{z}_i,\tilde{\mathbf{z}}_j),
\end{equation}
with RBF kernel
\begin{equation}
    k(\mathbf{x},\mathbf{y})
    =
    \exp\left(
        -\frac{\|\mathbf{x}-\mathbf{y}\|_2^2}{2\sigma^2}
    \right).
\end{equation}
The quadrotor latent-dynamics scripts use $\sigma=2.0$ for this kernel.  As in
the main paper, the Gaussian prior is only a convenient sampling space, not a
physical wind or payload model.

\paragraph{MMD schedule.}
The MMD weight is annealed:
\begin{equation}
    \lambda_{\mathrm{mmd}}(e)
    =
    \lambda_{\mathrm{mmd}}^{\max}
    \min\left(
        1,
        \max\left(0,\frac{e-E_0}{E_{\mathrm{ramp}}}\right)
    \right),
\end{equation}
where $e$ is the latent-dynamics update index, $E_0$ is the dynamics-only warmup,
and $E_{\mathrm{ramp}}$ is the ramp length.  This schedule addresses the failure
mode described in the main text.
If prior alignment is applied before the dynamics loss has separated incompatible conditions, the latent space can
collapse. 
If MMD is removed, encoded latents may be predictive on the buffer
but unsafe to sample during policy learning.

\subsection{Policy Learning and Deployment}
\label{app:policy_learning}

During differentiable policy optimization, the dynamics model and encoder are
frozen.  At the start of each simulated rollout, a latent is sampled from the
prior and held fixed:
\begin{equation}
    \mathbf{z}^{(j)} \sim \mathcal{N}(\mathbf{0},\mathbf{I})
    \quad \text{for environment } j.
\end{equation}
The same latent conditions both the residual model and the policy.  The policy
is optimized by backpropagation through time over parallel rollouts.  We write
\begin{equation}
    \bar{\mathbf{o}}_t
    =
    \mathrm{norm}_{o}\!\left(
        O_{\mathrm{task}}(\mathbf{s}_t,\mathbf{r}_t,\mathbf{a}_{t-K:t-1})
    \right)
\end{equation}
for the normalized task observation given to the policy, where
$O_{\mathrm{task}}$ selects the task-specific state, reference, and action-history
features and $\mathrm{norm}_{o}$ applies the observation normalization used for
policy training.  The latent-dynamics encoder input remains the separate
context window $\mathcal{H}_{t-C:t}$.  The BPTT objective is
\begin{equation}
    \max_{\theta}
    \mathbb{E}_{\mathbf{z},\mathbf{s}_0}
    \left[
        \sum_{t=0}^{H-1}
        \gamma^t
        r(\mathbf{s}_t,\pi_\theta(\bar{\mathbf{o}}_t,\mathbf{z}),\mathbf{s}_{t+1})
    \right].
\end{equation}
At deployment, the sampled latent is replaced by the live encoder output
$\mathbf{z}_t=E_\phi(\mathcal{H}_{t-C:t})$. 
Thus, the policy interface is the
same in simulation and on the robot, but the source of the latent changes from
prior samples during policy training to inferred recent interaction during
execution.

\end{document}